\documentclass[conference]{IEEEtran}
\IEEEoverridecommandlockouts

\usepackage{cite}
\usepackage{amsmath,amssymb,amsfonts, mathtools, array}

\usepackage{graphicx}
\usepackage{textcomp}
\usepackage{xcolor}
\usepackage{algorithm}
\usepackage{algpseudocode}
\usepackage{caption}
\usepackage{subcaption}
\usepackage{tabularx}
\usepackage{float}
\usepackage{lipsum,mathtools}
\usepackage{multirow}
\usepackage{dirtytalk}

\usepackage{url}

\usepackage{amssymb, nccmath}
\usepackage[numbers, sort&compress]{natbib}
\def\BibTeX{{\rm B\kern-.05em{\sc i\kern-.025em b}\kern-.08em
    T\kern-.1667em\lower.7ex\hbox{E}\kern-.125emX}}
    \IEEEaftertitletext{\vspace{-1\baselineskip}}
    \usepackage[left=0.58in, right=0.58in, top=0.71in, bottom=1.05in]{geometry} 
\begin{document}
\title{Semi-Supervised Federated Multi-Label Feature Selection with Fuzzy Information Measures}
\author{
    Afsaneh Mahanipour\textsuperscript{1},
    Hana Khamfroush\textsuperscript{1}\\[0.5ex]
    \textsuperscript{1}Department of Computer Science, University of Kentucky, Lexington, KY, USA\\
    ama654@uky.edu, khamfroush@cs.uky.edu
}
\maketitle
\begin{abstract}
Multi-label feature selection (FS) reduces the dimensionality of multi-label data by removing irrelevant, noisy, and redundant features, thereby boosting the performance of multi-label learning models. However, existing methods typically require centralized data, which makes them unsuitable for distributed and federated environments where each device/client holds its own local dataset. Additionally, federated methods often assume that clients have labeled data, which is unrealistic in cases where clients lack the expertise or resources to label task-specific data. To address these challenges, we propose a Semi-Supervised Federated Multi-Label Feature Selection method, called SSFMLFS, where clients hold only unlabeled data, while the server has limited labeled data. SSFMLFS adapts fuzzy information theory to a federated setting, where clients compute fuzzy similarity matrices and transmit them to the server, which then calculates feature redundancy and feature-label relevancy degrees. A feature graph is constructed by modeling features as vertices, assigning relevancy and redundancy degrees as vertex weights and edge weights, respectively. PageRank is then applied to rank the features by importance. Extensive experiments on five real-world datasets from various domains, including biology, images, music, and text, demonstrate that SSFMLFS outperforms other federated and centralized supervised and semi-supervised approaches in terms of three different evaluation metrics in non-IID data distribution setting.
\end{abstract}


\section{Introduction}
With the rise of technologies developed to realize smart cities, such as smart healthcare, Internet-of-Things (IoT), and intelligent transportation, we are now in the era of big data, with vast amounts of high-dimensional data generated daily. For instance, gene expression data can have 5,000 to 50,000 features. This data may include irrelevant, noisy, or redundant features, which increase the complexity and execution time of learning models, ultimately affecting their performance \cite{zebari2020comprehensive}.

To address these issues, data pre-processing methods, particularly feature selection (FS) techniques, are effective. Unlike dimensionality reduction methods like PCA, FS retains the original features by selecting the most relevant ones. This approach reduces data dimensions, lowers computational costs, and improves model performance. In domains like image annotation, text categorization, and gene prediction, where instances are associated with multiple labels, FS becomes even more crucial to manage high dimensionality effectively \cite{zebari2020comprehensive}.

Processing large amounts of data collected by different devices/clients within distributed environments is crucial for gaining valuable insights about the problem. While data was traditionally sent to cloud servers, the demand for real-time responses and privacy concerns has driven a shift to local or edge processing \cite{nishio2019client}. However, sending raw local data to an edge server increases communication costs, privacy risks, and delays. Moreover, using raw data locally in distributed, privacy-preserving models like federated learning (FL) increases complexity cost and execution time.

Many studies focus on supervised learning, assuming clients have fully labeled data. In reality, clients often lack the expertise or resources to label their data, making this process costly and time-consuming. For instance, a healthcare system or image/text classification system might have a server with limited labeled data from experts, while numerous clients hold large amounts of unlabeled data \cite{diao2022semifl}.

In a distributed environment where multiple clients hold unlabeled data and a server contains multi-label data, selecting informative features requires a collaborative semi-supervised federated feature selection (FFS) approach. Existing centralized multi-label FS methods struggle in these settings, often leading to bias and inaccurate results when applied independently on client data. To address this, we introduce the first semi-supervised federated multi-label FS method. This approach selects informative features from unlabeled client data using fuzzy complementary joint entropy and mutual information, while the server evaluates feature-label relevance using k-nearest neighbor fuzzy dependency. A feature graph is then constructed with features as vertices, using the weighted PageRank algorithm. Vertex weights reflect feature relevance, and edge weights capture inter-feature redundancy across clients. PageRank then determines the importance of features. This method efficiently identifies key features, reduces data size, accelerates learning models, and minimizes communication costs without losing information. The key innovations of this approach include:
\begin{itemize}
\item Introducing the first semi-supervised federated multi-label feature selection method (SSFMLFS) by integrating a semi-supervised approach into the FFS process.
\item Developing a federated adaptation of fuzzy information theory for feature evaluation in multi-label datasets.
\item Conducting extensive experiments on five real-world datasets from various domains, showing that SSFMLFS outperforms existing supervised and semi-supervised state-of-the-art multi-label feature selection methods.
\end{itemize}

\section{Related Works}
\vspace{-1mm}
Most previous research on multi-label FS has been centralized. To our knowledge, there is limited work on FFS and no research on semi-supervised federated multi-label FS. \textbf{(a) Centralized Multi-label Feature Selection:}
Multi-label feature selection methods are mainly classified into problem transformation and algorithm adaptation approaches. Problem transformation methods, such as Binary Relevance (BR), Pruned Problem Transformation (PPT), Label Powerset (LP), and Entropy-based Label Assignment (ELA), convert multi-label data into single-label data for easier feature selection. However, they face challenges like disrupting label correlations (BR) or struggling with imbalanced classes (LP) \cite{kashef2018multilabel}.

Algorithm adaptation methods extend supervised and semi-supervised FS techniques to handle multi-label datasets effectively. For instance, SCFC \cite{xu2018semi} leverages probabilistic neighborhood similarities for feature correlation. SFAM \cite{lv2021semi} integrates adaptive global structure learning with manifold learning. MLFS-NRS \cite{sun2021feature} employs the Fisher score and neighborhood rough sets for supervised FS, while ant colony optimization in MLACO \cite{paniri2020mlaco} balances feature-label relevancy and redundancy. MGFS \cite{hashemi2020mgfs} converts the FS problem into a graph and ranks features using PageRank, whereas GRMFS \cite{yin2023robust} uses fuzzy rough sets and weighted graph to measure feature-label uncertainty and improve feature selection. However, centralized methods are not suitable for distributed environments. \textbf{(b) Federated Feature Selection:}
A few supervised FFS methods are designed for single-label and multi-label datasets. These methods, inspired by federated learning (FL), are categorized into vertical FFS, which handles clients with identical instances but different feature sets, and horizontal FFS, where clients have the same features but different instances \cite{mahanipour2023wrapper}. For example, FMLFS \cite{mahanipour2024fmlfs} uses mutual information for feature-label relevance and correlation distance for redundancy, while FedCMFS \cite{song2024causal} introduces a federated causal approach for multi-label FS.
\vspace{-2mm}
\section{Preliminaries}
\textbf{(a) Fuzzy relation:}
Let \(\textit{U}=\{x_1, x_2, ..., x_n\}\) be a finite set of objects. If \(\mathcal A\) maps elements of \(\textit{U}\) to values in \([0,1]\), denoted as \(\mathcal A:\textit{U}\rightarrow[0,1]\), then \(\mathcal{A}=(\mathcal A(x_1), \mathcal A(x_2), ..., \mathcal A(x_n))\) is a fuzzy set on \(\textit{U}\). The membership function of each \(x\in \textit{U}\) is \(\mathcal A(x)\). 

\textbf{Definition 1.} A fuzzy relation \(\mathcal{R}\) on a set \(\textit{U}\) is defined as \(\mathcal{R}:\textit{U} \times \textit{U} \rightarrow [0,1]\). \(\mathcal{R} \in \textit{F}(\textit{U} \times \textit{U})\) where \(\textit{F}(\textit{U} \times \textit{U})\) is the set of all fuzzy relations on \(\textit{U}\). For any pair \((x,y) \in \textit{U} \times \textit{U}\), the membership degree \(\mathcal R(x,y)\) represents the extent to which \(x\) and \(y\) are related under the relation \(\mathcal R\). If \(\mathcal{R}\) meets the following conditions for any \(x, y, z \in \textit{U}\), it is considered a fuzzy equivalence relation:
\begin{enumerate}
\item Reflexivity: \(\mathcal R(x,x)=1\);
\item Symmetry: \(\mathcal R(x,y)=\mathcal R(y,x)\);
\item Transitivity: \(\mathcal R(x,z)\geq sup_{y \in U}\ min(\mathcal R(x,y), \mathcal R(y,z))\),
\end{enumerate}
If \(\mathcal R\) satisfies only conditions (1) and (2), it is referred to as a fuzzy similarity relation on \(\textit{U}\).

\textbf{(b) Fuzzy rough sets:}
The fuzzy rough set (FRS) model, introduced in \cite{dubois1990rough}, is defined as follows:

\textbf{Definition 2.} For a fuzzy equivalence relation \(\mathcal R\) on \(\textit{U}\) and any \(\mathcal{X} \in \textit{F(U)}\), the lower approximation \(\underline{\mathcal{R}}\mathcal{X}\) and upper approximation \(\overline{\mathcal{R}}\mathcal{X}\) of \(\mathcal{X}\) are fuzzy sets on \(\textit{U}\) with membership functions given by:
\vspace{-2mm}
\begin{equation}\label{my_forth_eqn}
\underline{\mathcal{R}}\mathcal{X}(x) = \inf_{y \in U}\ max \{1-\mathcal R(x, y), \mathcal X (y)\},
\end{equation}
\vspace{-2mm}
\begin{equation}\label{my_forth_eqn}
\overline{\mathcal{R}}\mathcal{X}(x) = \sup_{y \in U}\ min \{\mathcal R(x, y), \mathcal X (y)\}.
\end{equation}

\section{Proposed Method}
\subsection{System Overview}
\begin{figure}
\includegraphics[width=0.5\linewidth]{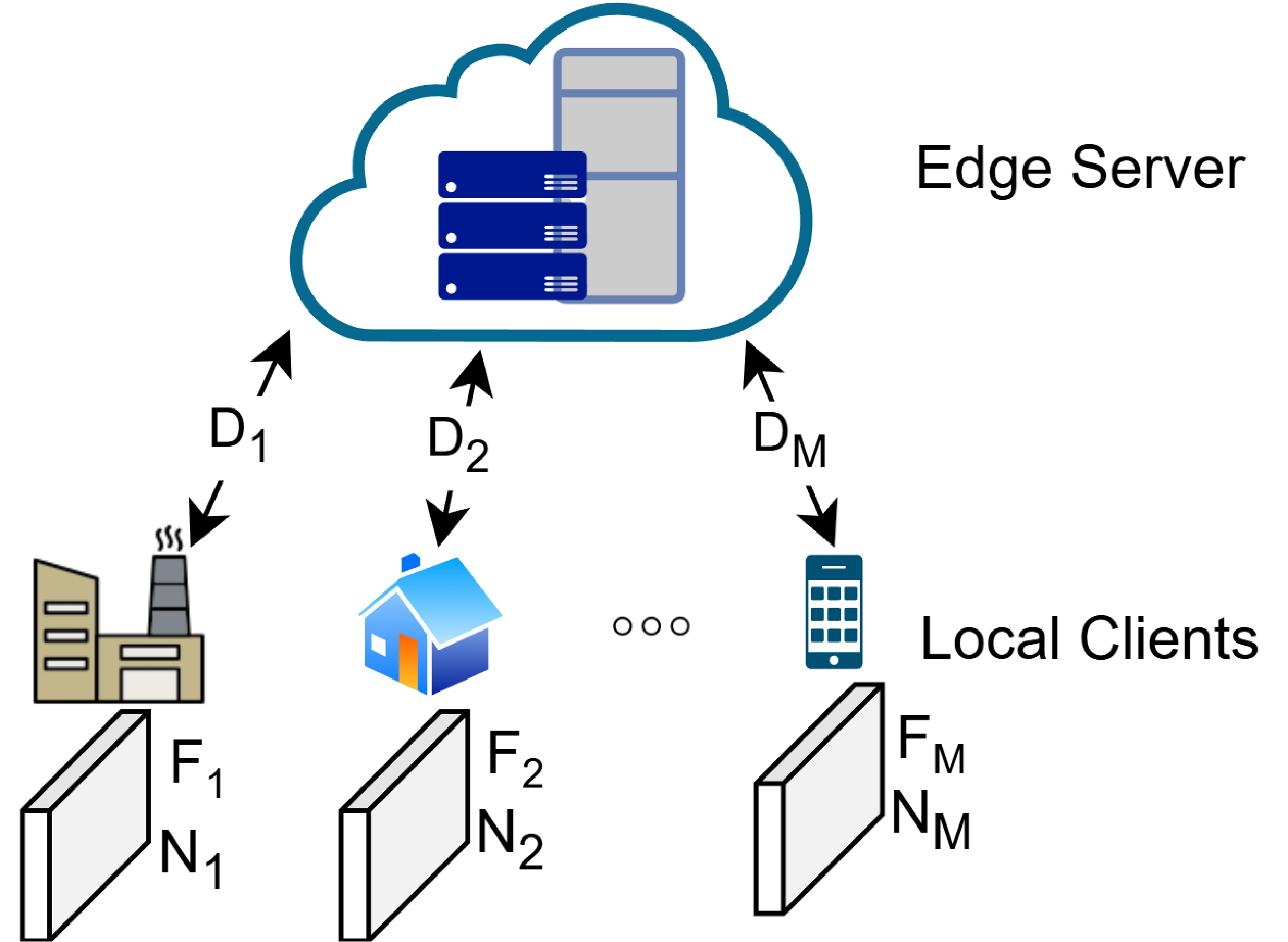}
\centering
\caption{Architecture overview of the two-layer system.}
\label{fig1}
\end{figure}

We examine a two-tier horizontal FFS framework, as depicted in Fig. \ref{fig1}, where multiple clients collect unlabeled data \(\mathcal Z\), and an edge server with labeled data \(\mathcal S\) coordinates feature selection. The system includes \(M\) clients (\(C_m, m=\{1, ..., M\}\)) and one edge server, scalable with additional servers, and requires at least two clients to avoid centralization. Client datasets \(\mathcal Z=\mathbb R ^{n\times d}\) contain \(d\)-dimensional feature vectors \(x_i=(x_{i1}, x_{i2}, ..., x_{id})\). The server’s labeled dataset \(\mathcal S=\mathbb R ^{s\times d}\), represented as \(\mathcal S=\{(x_i, y_i)\}_{i=1}^s\), includes \(s\) instances with \(d\)-dimensional feature vectors and \(L\)-dimensional binary labels \(y_i=(y_{i1}, y_{i2}, ..., y_{iL})\), where \(y_{il}=1\) if an instance has label \(y_{l}\), and 0 otherwise. The goal is to collaboratively select informative features. 

\vspace{-2mm}
\subsection{Proposed Algorithm}
The proposed method, SSFMLFS (Semi-Supervised Federated Multi-Label Feature Selection), integrates semi-supervised and federated learning to select informative features across client datasets. It uniquely adapts fuzzy complementary entropy and related uncertainty metrics to federated settings, minimizing data transfer while enhancing speed and proximity to data sources. SSFMLFS evaluates features based on maximum information and minimum redundancy.

The fuzzy information system for clients, denoted as \(FIS= 〈U, F〉\), characterizes a system where \(\textit{U}=\{x_1, x_2, ..., x_n\}\) represents the data instances and \(\textit{F}=\{f_1, f_2, ..., f_d\}\) denotes the feature set. For any feature $\textit{\(f_p\)} \in \textit{F}$, \(\mathcal R_{f_p}\) is a fuzzy similarity relation on \(\textit{U}\), and $M(\mathcal R_{f_p})=(r_{ij}^{f_p})_{n \times n}$ denotes the fuzzy similarity relation matrix. To prevent varying feature value ranges from affecting algorithm performance, the original features are normalized. In this study, normalization scales feature values to a range of [0, 1]. Then, in each client, the standard deviation of each feature \(f_p\) is calculated as: \(std(f_p)=\sqrt{\frac{\sum_{i=1} ^n (f_p(x_i)-f_p(\overline{x}))^2}{n-1}}\), where \(f_p(\overline{x})\) is the mean of all instances in the local dataset for \(f_p\). The calculated standard deviation of all features are then sent from the clients to the edge server.

The server aggregates the standard deviations received from clients and computes the global standard deviation for each feature (\(std_g(f_p)\)). It then broadcasts these values to the clients. Using the global standard deviation, each client calculates the adaptive fuzzy radius \(\varepsilon_{f_p}=\frac{std_g(f_p)}{\lambda}\), where \(\lambda\in [0.4,2]\) adjusts the radius. Clients subsequently compute the fuzzy similarity relation matrix for each feature, determining the similarity \(r_{ij}^{f_p}\) between data instances \(x_i\) and \(x_j\) based on the method in \cite{wan2021interactive}.
\vspace{-2mm}
\begin{equation}\label{my_forth_eqn}
\small
\hfill r_{ij}^{f_p}=
\begin{cases}
1-|f_p(x_i)-f_p(x_j)|, \ if \ |f_p(x_i)-f_p(x_j)|\leqslant \varepsilon_{f_p}\\
0, \quad \quad \quad \quad \quad \quad \quad \quad \quad  if \ |f_p(x_i)-f_p(x_j)| > \varepsilon_{f_p}
\end{cases}
\end{equation}

Then, the fuzzy similarity relation matrices are transmitted to the server, where they are aggregated to obtain global fuzzy similarity relation matrices for all features. The server then evaluates features by calculating fuzzy complementary joint entropy and mutual information, maximizing information and minimizing redundancy, using FRS theory:
\textbf{Definition 3.} The Fuzzy complementary entropy for any feature \(f_p\) is defined as:
\vspace{-2mm}
\begin{equation}\label{my_forth_eqn}
CE(f_p)=\frac{1}{|U|} \sum_{i=1}^n \left(1-\frac{|[x_i]_{f_p}|}{|U|} \right).
\end{equation}

\noindent where \(|[x_i]_{f_p}|=\sum_{j=1}^n r_{ij}^{f_p}\). Fuzzy complementary entropy, an uncertainty measure for fuzzy information systems, is first defined for a single relation and then extended to multiple relations as fuzzy complementary joint entropy.

\textbf{Definition 4.} The fuzzy complementary joint entropy of \(f_p\) and \(f_b\) is defined as:
\begin{equation}\label{my_forth_eqn}
CJE(f_p, f_b)=\frac{1}{|U|} \sum_{i=1}^n \left(1-\frac{|[x_i]_{f_p} \cap [x_i]_{f_b}|}{|U|} \right).
\end{equation}

\textbf{Definition 5.} The fuzzy complementary conditional entropy of \(f_p\) on \(f_b\) is defined as:
\vspace{-1mm}
\begin{equation}\label{my_forth_eqn}
CCE(f_p|f_b)=\frac{1}{|U|} \sum_{i=1}^n \left(\frac{|[x_i]_{f_b}|}{|U|} - \frac{|[x_i]_{f_p} \cap [x_i]_{f_b}|}{|U|} \right).
\end{equation}

\textbf{Definition 6.} The fuzzy complementary mutual information between \(f_p\) and \(f_b\) is defined as:
\vspace{-1mm}
\begin{equation}\label{my_forth_eqn}
\begin{split}
&CMI(f_p; f_b)= \\
&\frac{1}{|U|} \sum_{i=1}^n \left(1 - \frac{|[x_i]_{f_b}| + |[x_i]_{f_p}| - |[x_i]_{f_p} \cap [x_i]_{f_b}|}{|U|} \right).
\end{split}
\end{equation}

After calculating fuzzy complementary joint entropy and mutual information between features, we calculate a correlation distance to maximize information and minimize redundancy by subtracting fuzzy complementary mutual information from fuzzy complementary joint entropy.

Now, using the labeled dataset on the edge server, we evaluate the features based on their relation to all labels. For this purpose, we calculate the k-nearest neighbor fuzzy dependency degree \cite{yin2023robust} as follows to determine the feature-label relevancy.

\textbf{Definition 7.} Given a multi-label fuzzy information system \(MFIS= 〈U, F \cup Y〉\), where \(F = \{f_1, f_2, . . . , f_d\}\) denotes the feature set and \(Y = \{y_1, y_2, \cdots , y_L\}\) denotes the label set, the fuzzy positive region of \(Y\) with respect to feature \(f_p \in F\) is denoted as:
\vspace{-2mm}
\begin{equation}\label{my_forth_eqn}
POS_{f_p}(Y)(x_i)=\bigcup_{Y_l} \underline{\mathcal R_{f_p}}(Y_l)(x_i).
\end{equation}

\textbf{Definition 8.} The fuzzy similarity between instances \(x_i\) and \(x_j\) under label set \(Y\) is defined as:
\begin{equation}\label{my_forth_eqn}
\begin{split}
&Sim(x_i, x_j) = \\
&\frac{\sum_{t=1}^L (\widetilde{Y}^t(x_i)- \overline{\widetilde{Y}(x_i)}) (\widetilde{Y}^t(x_j)-\overline{\widetilde{Y}(x_j)})}{\sqrt{\sum_{t=1}^L (\widetilde{Y}^t(x_i)- \overline{\widetilde{Y}(x_i)}) ^2} \sqrt{\sum_{t=1}^L (\widetilde{Y}^t(x_j)- \overline{\widetilde{Y}(x_j)}) ^2}}.
\end{split}
\end{equation}

\noindent where \(\widetilde{Y}^t(x_i)\) and \(\widetilde{Y}^t(x_j)\) are the fuzzy decisions of samples \(x_i\) and \(x_j\) on label \(y_t\), respectively. \(\overline{\widetilde{Y}(x_i)}\) represents the mean of the fuzzy decisions of \(x_i\) on the label set \(Y\). Then, k nearest relevant and irrelevant instances are selected as the soft same class sample set \(ST(x_i)\) and different class sample set \(DT(x_i)\), respectively.

\begin{algorithm}[tb]
\caption{Pseudocode of the proposed semi-supervised federated multi-label FS (SSFMLFS)}
\label{alg:algorithm}
\textbf{Input}: $M$ Clients with their unlabeled data \(\mathcal Z\), and labeled data on the server \(\mathcal S\), \(\lambda=1.2 \ \cite{yin2023robust}\).\\
\textbf{Output}: Selected feature subset
\begin{algorithmic}[1] 
\State\noindent\textbf{Client}
\For{$f_p\in F$}
\State Calculate standard deviation;
\EndFor
\State\noindent\textbf{Server}
\State Aggregate the standard deviations received from clients
\State\noindent\textbf{Client}
\For{$f_p\in F$}
\State Calculate fuzzy similarity matrix \(M(\mathcal R_{f_p})\) (Eq. 3);
\EndFor
\State\noindent\textbf{Server}
\State Aggregate fuzzy similarity matrices
\For{$f_p\in F$}
    \For{$f_b\in F$}
        \State Compute fuzzy joint entropy \(CJE(f_p, f_b)\) (Eq. 5);
        \State Compute fuzzy mutual information \(CMI(f_p;f_b)\) (Eq. 7);
    \EndFor
\EndFor
\State \% The initial weights for edges in the graph \%
\State Calculate correlation distance\(\ = CJE - CMI\);
\State \% The initial weights for vertices in the graph \%
\State Calculate fuzzy similarity under the label set \(Y\) (Eq. 9);
\For{$f_p \in F$}
    \State Compute k-nearest neighbor dependency \(D_{f_p}^\lambda(Y)\) (Eq. 12);
\EndFor
\State Apply weighted PageRank (Eq. 13) to score features;
\State Sort feature scores in descending order;
\State Select the top features as specified by the user.
\end{algorithmic}
\end{algorithm}

\textbf{Definition 9.} The k-nearest neighbor fuzzy lower and upper approximations are defined as follows:
\vspace{-1mm}
\begin{equation}\label{my_forth_eqn}
\underline{\mathcal R_{f_p}^{\lambda,k}}(Y)(x_i)=\frac{1}{k}\sum_{j=1}^k (1-\mathcal R_{f_p}^\lambda (x_i, x_j)), \; x_j \in DT(x_i),
\end{equation}
\vspace{-1mm}
\begin{equation}\label{my_forth_eqn}
\overline{\mathcal R_{f_p}^{\lambda,k}}(Y)(x_i)=\frac{1}{k}\sum_{j=1}^k \mathcal R_{f_p}^\lambda (x_i, x_j), \; x_j \in ST(x_i).
\end{equation}

\textbf{Definition 10.} The k-nearest neighbor fuzzy dependency degree of \(Y\) to feature \(f_p\), \(D_{f_p}^\lambda(Y)\), is defined as follows:
\vspace{-1mm}
\begin{equation}\label{my_forth_eqn}
D_{f_p}^\lambda(Y)=\frac{\sum_{x_i \in U}POS_{f_p}^{\lambda,k}(Y)(x_i)}{|U|}.
\end{equation}

\noindent where \(POS_{f_p}^{\lambda,k}(Y)(x_i)=\underline{\mathcal R_{f_p}^{\lambda,k}}(Y)(x_i)\) represents the k-nearest neighbor fuzzy positive region. This fuzzy dependency degree evaluates the feature-label relevance degree.

\begin{table}[tb]
\caption{Details of the multi-label benchmark datasets.}
\vspace{-2mm}
\begin{center}
\resizebox{\columnwidth}{!}{%
\begin{tabular}{c|c|c|c|c}
\hline
Dataset & Instances & Features & Labels & Domain \\
\hline
Cal500 & 502 & 68 & 174 & Music\\
Corel5k & 5000 & 499 & 374 & Images\\
Emotions & 593 & 72 & 6 & Music\\
Enron & 1702 & 1001 & 53 & Text\\
Yeast & 2417 & 103 & 14 & Biology\\
\hline
\end{tabular}
}
\label{table1}
\end{center}
\end{table}

After evaluating the correlation distances between features and the relevancy between features and labels, we adopt a weighted PageRank algorithm \cite{luo2016ensemble} to traverse a feature graph and select the most informative features based on the weights of the edges and vertices. In this graph, features are considered as the vertices, the relevance between features and labels serves as the initial weights of the vertices, and the correlation distance between features serves as the initial weights of the edges. The PageRank algorithm then assigns scores (\(G_{f_i}\)) to the vertices iteratively, with features having higher scores considered more informative. The PageRank formula is as follows \cite{luo2016ensemble}:
\begin{equation}\label{my_forth_eqn}
G_{f_i} = (1-\zeta)W_{f_i} + \zeta \sum_{f_j \in B(f_i)}\frac{G_{f_j} W_{u_{ij}}}{\sum_{f_z \in B(f_j)}W_{u_{jz}}}.
\end{equation}

\noindent where \(\zeta\) is the probability of accessing the current feature, usually set to \(0.85\) \cite{yin2023robust}; \(W_{f_i}\) is the initial weight of feature \(f_i\), and \(W_{u_{ij}}\) is the weight of the edge between features \(f_i\) and \(f_j\). Also, \(B(f_i)\) is the set of features connected to feature \(f_i\). The features' scores are then ranked in descending order, and the desired number of best features are selected. The pseudocode for the SSFMLFS method is provided in Algorithm 1.

\section{Experimental Results}
In this section, we evaluate the proposed method against supervised FFS methods and centralized supervised and semi-supervised FS methods. Clients use SSFMLFS to select informative features and then send the smaller datasets to the edge server for comparison with other multi-label FS methods using the MLKNN classifier.

\begin{table}[htbp]
\caption{AP for SSFMLFS vs. ten supervised FS methods.}
\vspace{-2mm}
\label{table2}
\begin{center}
\begin{tabular}{l c c c c c}
\hline
\multirow{2}{*}{Algorithms} & \multicolumn{5}{c}{AP ($\uparrow$)} \\ 
\cline{2-6}
 & Cal500 & Corel5k & Emotions & Enron & Yeast \\
\hline
D2F & 0.4724 & 0.2217 & 0.6790 & 0.6556 & 0.7256 \\
RFNMIFS & 0.4839 & 0.2333 & 0.6702 & 0.5468 & 0.7375 \\
FIMF & 0.4774 & 0.2234 & 0.7122 & 0.6533 & 0.7414 \\
MLFS-NRS & 0.4709 & 0.2311 & 0.6866 & 0.5827 & 0.7449 \\
MFSDNA & 0.4758 & 0.2242 & 0.7199 & 0.5293 & 0.7508 \\
MLNRS & 0.4729 & 0.2081 & 0.7233 & 0.5250 & 0.7489 \\
BMFS & 0.4871 & 0.2107 & 0.6703 & 0.5578 & 0.7220 \\
MGFS & 0.4883 & 0.2309 & 0.7165 & 0.6578 & 0.7448 \\
MLACO & 0.4976 & 0.2176 & 0.6552 & 0.6498 & 0.7463 \\
GRMFS & 0.5012 & 0.2360 & 0.7309 & \textbf{0.6698} & 0.7582 \\
SSFMLFS & \textbf{0.5119} & \textbf{0.2409} & \textbf{0.7749} & 0.6687 & \textbf{0.7792} \\
\hline
\end{tabular}
\end{center}
\end{table}

\noindent\textbf{(a) Datasets:}
This study selects five real-world multi-label datasets from diverse domains, including biology, image, text, and music, sourced from the Mulan\footnote[1]{\url{https://mulan.sourceforge.net/datasets.html}} repository. The datasets vary in the number of instances, features, and labels. Table \ref{table1} lists the characteristics of these datasets.

\begin{figure*}
\includegraphics[width=0.9\textwidth]{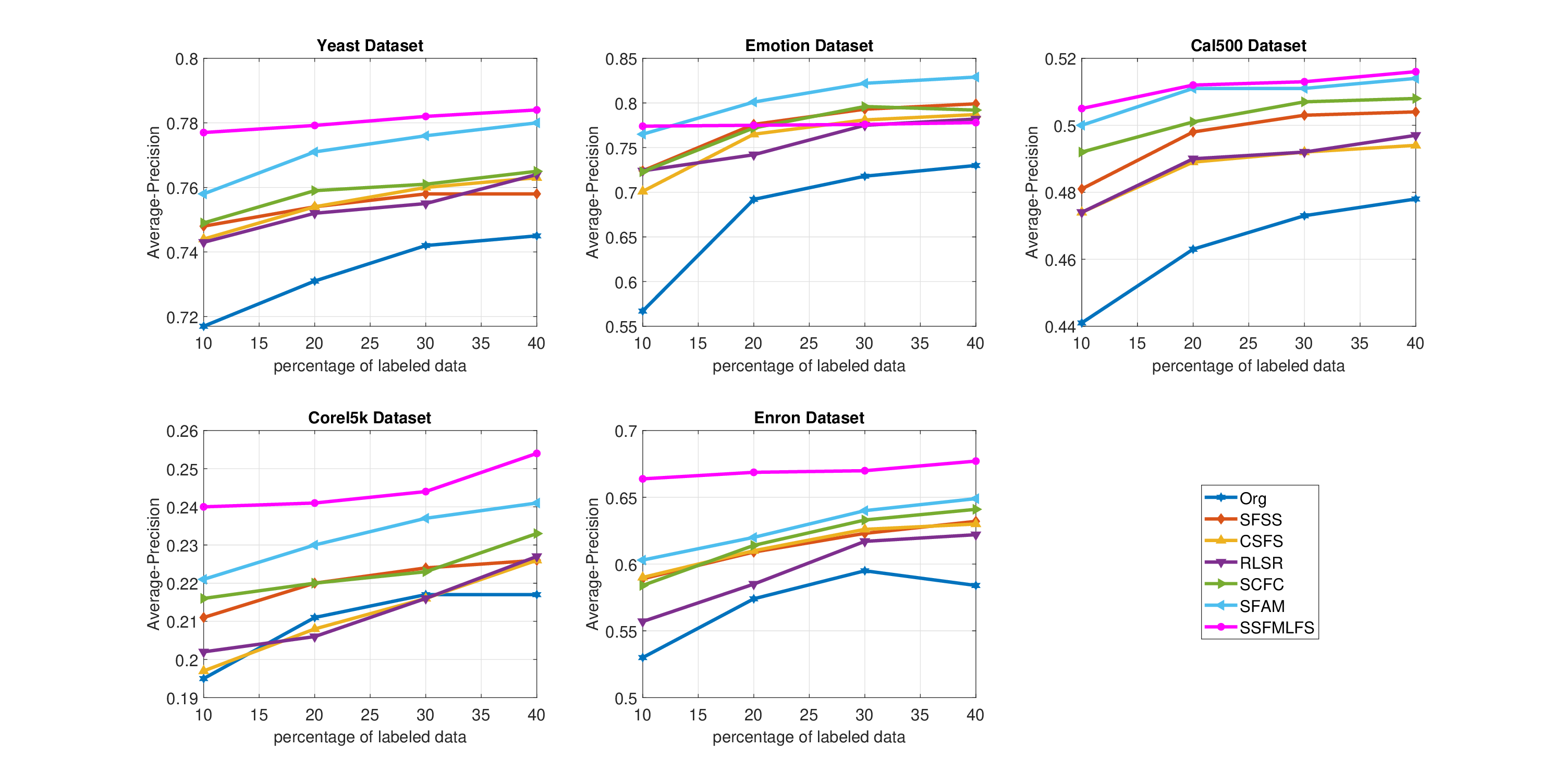}
\centering
\vspace{-2mm}
\caption{Performance comparison across all datasets with varying labeled data percentages (\(10\%, 20\%, 30\%, 40\%\)).}
\label{fig2}
\end{figure*}

\begin{table}[htbp]
\caption{CV for SSFMLFS vs. ten supervised FS methods.}
\vspace{-2mm}
\label{table3}
\begin{center}
\begin{tabular}{l c c c c c}
\hline
\multirow{2}{*}{Algorithms} & \multicolumn{5}{c}{CV ($\downarrow$)} \\ 
\cline{2-6}
 & Cal500 & Corel5k & Emotions & Enron & Yeast \\
\hline
D2F & 133.5412 & 117.1810 & 2.400 & 12.7824 & 6.6704 \\
RFNMIFS & 133.0102 & 115.7105 & 2.3125 & 13.5924 & 6.5857 \\
FIMF & 132.2130 & 116.9134 & 2.2566 & 12.6118 & 6.5682 \\
MLFS-NRS & 135.1731 & 116.6408 & 2.1429 & 13.5015 & 6.4838 \\
MFSDNA & 135.1193 & 118.3408 & 2.1261 & 15.2815 & 6.4903 \\
MLNRS & 133.4587 & 119.8625 & 2.1739 & 15.0749 & 6.4869 \\
BMFS & 130.7996 & 118.5782 & 2.5781 & 14.2853 & 6.8413 \\
MGFS & 131.2896 & 116.699 & 2.2143 & 12.8417 & 6.4794 \\
MLACO & 130.4024 & 119.5255 & 2.5072 & 12.8798 & 6.5037 \\
GRMFS & 129.0146 & 112.143 & 2.1243 & \textbf{12.4176} & 6.4707 \\
SSFMLFS & \textbf{128.3000} & \textbf{89.9625} & \textbf{2.0420} & 12.8647 & \textbf{6.1242} \\
\hline
\end{tabular}
\end{center}
\end{table}

\vspace{-4mm}

\noindent\textbf{(b) Baselines:}
We compare the proposed method with five semi-supervised and ten supervised centralized multi-label FS methods, as well as two supervised multi-label FFS methods. The methods are as follows: \textbf{Semi-supervised methods} \cite{lv2021semi}\textbf{:} SFSS, CSFS, RLSR, SCFC, and SFAM. \textbf{Supervised methods} \cite{yin2023robust}\textbf{:} Org uses all the original features in the experiments. D2F, FSNM, RFNMIFS, FIMF, MLFS-NRS, MFSDNA, MLFRS, BMFS, MGFS, MLACO, GRMFS, FMLFS \cite{mahanipour2024fmlfs}, and FedCMFS \cite{song2024causal}.
The experiments are run in an integrated development environment of MATLAB R2023b on a computer with Windows 11 with Intel(R) Core(TM) i7-8700 CPU at 3.20 GHz and 32.0 GB memory. 

\noindent\textbf{(c) Evaluation Metrics:}
Three evaluation metrics are used to assess the performance of the proposed method: average precision (AP), coverage (CV), and ranking loss (RL). Denoting a test set as (\(\mathcal T=\{(x_i, y_i)\}_{i=1}^n\)), where \(y_i\) and \(z_i\) are the actual and predicted label sets for \(x_i\), respectively, these metrics are as follows \cite{kashef2018multilabel}:

\begin{table}[htbp]
\caption{RL for SSFMLFS vs. ten supervised FS methods.}
\vspace{-2mm}
\label{table4}
\begin{center}
\begin{tabular}{l c c c c c}
\hline
\multirow{2}{*}{Algorithms} & \multicolumn{5}{c}{RL ($\downarrow$)} \\ 
\cline{2-6}
 & Cal500 & Corel5k & Emotions & Enron & Yeast \\
\hline
D2F & 0.1902 & 0.1389 & 0.2858 & 0.0887 & 0.1954 \\
RFNMIFS & 0.1894 & 0.1366 & 0.2977 & 0.1026 & 0.1871 \\
FIMF & 0.1941 & 0.1383 & 0.2663 & 0.0905 & 0.1845 \\
MLFS-NRS & 0.1904 & 0.1374 & 0.2576 & 0.0933 & 0.1777 \\
MFSDNA & 0.1940 & 0.1407 & 0.2654 & 0.1131 & 0.1772 \\
MLNRS & 0.1910 & 0.1428 & 0.2708 & 0.1143 & 0.1751 \\
BMFS & 0.1858 & 0.1411 & 0.3124 & 0.1050 & 0.1990 \\
MGFS & 0.1855 & 0.1376 & 0.2310 & 0.0912 & 0.1710 \\
MLACO & 0.1849 & 0.1442 & 0.2971 & 0.0879 & 0.1818 \\
GRMFS & 0.1804 & 0.1318 & 0.2215 & \textbf{0.0825} & 0.1710 \\
SSFMLFS & \textbf{0.1723} & \textbf{0.1051} & \textbf{0.2064} & 0.0847 & \textbf{0.1527} \\
\hline
\end{tabular}
\end{center}
\end{table}

\vspace{-1.5mm}
\begin{itemize}
\item AP calculates the average fraction of relevant labels ranked higher than a specific label.
\vspace{-2mm}
\begin{equation}\label{my_forth_eqn}
\small
AP = \frac{1}{n} \sum_{i=1}^n \frac{1}{|y_i|}
\sum_{\lambda \in y_i} \frac{|{\lambda^{\prime} \in y_i : rank(\lambda^{\prime}) \leq rank(\lambda)}|}{rank(\lambda)}
\end{equation}

\item CV indicates the number of steps needed for a learning algorithm to cover all true labels of an instance.
\vspace{-3mm}
\begin{equation}\label{my_forth_eqn}
CV = \frac{1}{n} \sum_{i=1}^n \max_{\lambda \in y_i} (rank(\lambda)) -1
\end{equation}
\vspace{-2mm}
\item RL assesses how often relevant labels are ranked lower than non-relevant labels.
\vspace{-1mm}
\begin{equation}\label{my_forth_eqn}
\begin{split}
&RL = \frac{1}{n} \sum_{i=1}^n \frac{1}{|y_i||\Bar{y_i}|} |{(\lambda_a,\lambda_b):rank(\lambda_a)> rank(\lambda_b),}\\
& {(\lambda_a,\lambda_b)\in y_i \times \Bar{y_i}}|
\end{split}
\end{equation}
\vspace{-1mm}
\noindent where \(\Bar{y_i}\) is the complement set of \(y_i\).

\end{itemize}

\noindent ($\downarrow$) indicates that \say{lower values are better}, while ($\uparrow$) denotes that \say{higher values are better} for each metric.

\noindent\textbf{(d) Parameter settings:}
For the experiments, we use MLKNN with \(k=10\) and a smoothing parameter of 1 as the classifier, which is a common choice for evaluating federated and centralized multi-label FS results. Additionally, in line with existing FFS methods, we involve 10 clients with non-IID data and apply the classifier after selecting the desired number of features. The final number of selected features for comparison is determined based on the method recommended in \cite{kashef2018multilabel}.

\begin{table}[htbp]
\caption{Comparison of FFS methods on the Yeast dataset.}
\label{table7}
\setlength{\tabcolsep}{2.3\tabcolsep}
\begin{center}
\begin{tabular}{l c c c }
\hline
\multirow{2}{*}{Algorithms} & \multicolumn{3}{c}{Yeast} \\ 
\cline{2-4}
 & AP ($\uparrow$) & CV ($\downarrow$) & RL ($\downarrow$)   \\
\hline
FMLFS & 0.7534 & 6.4199 & 0.1732   \\
FedCMFS & 0.7590 & 6.4531 & 0.1717   \\
SSFMLFS & \textbf{0.7792} & \textbf{6.1242} & \textbf{0.1527}   \\
\hline
\end{tabular}
\end{center}
\end{table}

\vspace{-1mm}
\noindent\textbf{(e) Results and Analysis:}
Fig. \ref{fig2} compares our federated FS method with various centralized semi-supervised multi-label FS approaches. The results show that SSFMLFS selects more informative features with fewer labeled instances than other centralized methods across all datasets except for \say{Emotions}. For instance, with only \(10\%\) of labeled instances in the \say{corel5k} and \say{Enron} datasets, our method, using less labeled data, matches or outperforms the performance of \say{SFAM}, the second-best method, which uses \(40\%\) labeled data. Table \ref{table2} to \ref{table4} present comparison results against ten supervised centralized multi-label FS methods, covering approaches based on graph, rough sets, and information theory, across all datasets using three evaluation metrics. The results for the proposed method are based on utilizing \(20\%\) of the labeled training data on the server. The findings reveal that incorporating both labeled and unlabeled data allows the proposed method to outperform others in nearly all metrics. For instance, the \say{Yeast} and \say{Emotions} datasets show \(2.1\%\) and \(4.4\%\) improvements in AP, respectively, compared to \say{GRMFS}, the second-best method. Additionally, Table \ref{table7} shows that SSFMLFS, with only \(20\%\) labeled data, outperforms two supervised FFS methods on the \say{Yeast} dataset.

\noindent\textbf{Communication Cost:} To calculate the communication cost before and after FS, the following formula is used. As shown in Table \ref{table8}, the results demonstrate that the communication cost decreased after FS due to the reduced number of features.

\begin{equation}\label{my_forth_eqn}
Communivation-Cost = \sum_{i=1}^M D_i \times N_i \times F_i \times b
\end{equation}

\noindent Here, \(M\) is the number of clients, \(D_i\) is the distance between client \(i\) and the edge server, \(N_i\) and \(F_i\) represent the number of instances and features, respectively, in the local dataset, and \(b\) indicates the number of bits per value (e.g., 32 for float). 

\begin{table}[htbp]
\caption{Communication Cost before and after FS.}
\vspace{-3mm}
\label{table8}
\setlength{\tabcolsep}{0.80\tabcolsep}
\begin{center}
\begin{tabular}{l c c }
\hline
\multirow{2}{*}{Datasets} & \multicolumn{2}{c}{Communivation Cost ($\downarrow$)} \\ 
\cline{2-3}
 & Before FS & After FS \\
\hline
Cal500 & \(\sum_{i=1}^M D_i \times 502 \times 68 \times b \) & \(\sum_{i=1}^M D_i \times 502 \times 27 \times b\) \\
Corel5k & \(\sum_{i=1}^M D_i \times 5000 \times 499 \times b \) & \(\sum_{i=1}^M D_i \times 5000 \times 150 \times b\) \\
Emotion & \(\sum_{i=1}^M D_i \times 593 \times 72 \times b \) & \(\sum_{i=1}^M D_i \times 593 \times 28 \times b\) \\
Enron & \(\sum_{i=1}^M D_i \times 1702 \times 1001 \times b \) & \(\sum_{i=1}^M D_i \times 1702 \times 100 \times b\) \\
Yeast & \(\sum_{i=1}^M D_i \times 2417 \times 103 \times b \) & \(\sum_{i=1}^M D_i \times 2417 \times 31 \times b\) \\
\hline
\end{tabular}
\end{center}
\end{table}

\vspace{-5mm}
\section{Conclusion and Future Works}

In this paper, we present SSFMLFS, a novel semi-supervised federated feature selection method for multi-label data. SSFMLFS adapts fuzzy mutual information and joint entropy into a federated setting to select informative features from unlabeled client data and limited labeled server data. Extensive experiments are conducted on 5 diverse datasets with four varying labeled data percentages. SSFMLFS outperforms other methods in 17 out of 20 comparisons against various semi-supervised centralized multi-label FS methods. Furthermore, compared to ten supervised centralized FS methods that have \(100\%\) labeled data, SSFMLFS, with only \(20\%\) labeled data, performs better across nearly all metrics on four datasets. For instance, on the \say{Emotions} dataset, it improves AP by \(4.4\%\) over the second-best method, \say{GRMFS}, and by \(20\%\) over \say{Org}. Finally, SSFMLFS demonstrates better performance compared to other supervised FFS methods. Future work will explore integrating encryption techniques and embedded FFS with semi-supervised FL in non-IID data distribution setting.
\vspace{-3mm}
\section*{Acknowledgement}
\vspace{-2mm}
This work is funded by career grant provided by the National Science Foundation (NSF) under the grant number 2340075.
\bibliographystyle{IEEEtranN}
\bibliography{aaai25}

\begin{thebibliography}{16}
\providecommand{\natexlab}[1]{#1}
\providecommand{\url}[1]{#1}
\csname url@samestyle\endcsname
\providecommand{\newblock}{\relax}
\providecommand{\bibinfo}[2]{#2}
\providecommand{\BIBentrySTDinterwordspacing}{\spaceskip=0pt\relax}
\providecommand{\BIBentryALTinterwordstretchfactor}{4}
\providecommand{\BIBentryALTinterwordspacing}{\spaceskip=\fontdimen2\font plus
\BIBentryALTinterwordstretchfactor\fontdimen3\font minus \fontdimen4\font\relax}
\providecommand{\BIBforeignlanguage}[2]{{%
\expandafter\ifx\csname l@#1\endcsname\relax
\typeout{** WARNING: IEEEtranN.bst: No hyphenation pattern has been}%
\typeout{** loaded for the language `#1'. Using the pattern for}%
\typeout{** the default language instead.}%
\else
\language=\csname l@#1\endcsname
\fi
#2}}
\providecommand{\BIBdecl}{\relax}
\BIBdecl

\bibitem[Zebari et~al.(2020)Zebari, Abdulazeez, Zeebaree, Zebari, and Saeed]{zebari2020comprehensive}
R.~Zebari, A.~Abdulazeez, D.~Zeebaree, D.~Zebari, and J.~Saeed, ``A comprehensive review of dimensionality reduction techniques for feature selection and feature extraction,'' \emph{Journal of Applied Science and Technology Trends}, vol.~1, no.~2, pp. 56--70, 2020.

\bibitem[Nishio and Yonetani(2019)]{nishio2019client}
T.~Nishio and R.~Yonetani, ``Client selection for federated learning with heterogeneous resources in mobile edge,'' in \emph{ICC 2019-2019 IEEE international conference on communications (ICC)}.\hskip 1em plus 0.5em minus 0.4em\relax IEEE, 2019, pp. 1--7.

\bibitem[Diao et~al.(2022)Diao, Ding, and Tarokh]{diao2022semifl}
E.~Diao, J.~Ding, and V.~Tarokh, ``Semifl: Semi-supervised federated learning for unlabeled clients with alternate training,'' \emph{Advances in Neural Information Processing Systems}, vol.~35, pp. 17\,871--17\,884, 2022.

\bibitem[Kashef et~al.(2018)Kashef, Nezamabadi-pour, and Nikpour]{kashef2018multilabel}
S.~Kashef, H.~Nezamabadi-pour, and B.~Nikpour, ``Multilabel feature selection: A comprehensive review and guiding experiments,'' \emph{Wiley Interdisciplinary Reviews: Data Mining and Knowledge Discovery}, vol.~8, no.~2, p. e1240, 2018.

\bibitem[Xu et~al.(2018)Xu, Wang, An, Wei, and Ruan]{xu2018semi}
Y.~Xu, J.~Wang, S.~An, J.~Wei, and J.~Ruan, ``Semi-supervised multi-label feature selection by preserving feature-label space consistency,'' in \emph{Proceedings of the 27th ACM international conference on information and knowledge management}, 2018, pp. 783--792.

\bibitem[Lv et~al.(2021)Lv, Shi, Wang, and Li]{lv2021semi}
S.~Lv, S.~Shi, H.~Wang, and F.~Li, ``Semi-supervised multi-label feature selection with adaptive structure learning and manifold learning,'' \emph{Knowledge-based systems}, vol. 214, p. 106757, 2021.

\bibitem[Sun et~al.(2021)Sun, Wang, Ding, Xu, and Lin]{sun2021feature}
L.~Sun, T.~Wang, W.~Ding, J.~Xu, and Y.~Lin, ``Feature selection using fisher score and multilabel neighborhood rough sets for multilabel classification,'' \emph{Information Sciences}, vol. 578, pp. 887--912, 2021.

\bibitem[Paniri et~al.(2020)Paniri, Dowlatshahi, and Nezamabadi-Pour]{paniri2020mlaco}
M.~Paniri, M.~B. Dowlatshahi, and H.~Nezamabadi-Pour, ``Mlaco: A multi-label feature selection algorithm based on ant colony optimization,'' \emph{Knowledge-Based Systems}, vol. 192, p. 105285, 2020.

\bibitem[Hashemi et~al.(2020)Hashemi, Dowlatshahi, and Nezamabadi-Pour]{hashemi2020mgfs}
A.~Hashemi, M.~B. Dowlatshahi, and H.~Nezamabadi-Pour, ``Mgfs: A multi-label graph-based feature selection algorithm via pagerank centrality,'' \emph{Expert Systems with Applications}, vol. 142, p. 113024, 2020.

\bibitem[Yin et~al.(2023)Yin, Chen, Yuan, Wan, Liu, Horng, and Li]{yin2023robust}
T.~Yin, H.~Chen, Z.~Yuan, J.~Wan, K.~Liu, S.-J. Horng, and T.~Li, ``A robust multilabel feature selection approach based on graph structure considering fuzzy dependency and feature interaction,'' \emph{IEEE Transactions on Fuzzy Systems}, vol.~31, no.~12, pp. 4516--4528, 2023.

\bibitem[Mahanipour and Khamfroush(2023)]{mahanipour2023wrapper}
A.~Mahanipour and H.~Khamfroush, ``Wrapper-based federated feature selection for iot environments,'' in \emph{2023 International Conference on Computing, Networking and Communications (ICNC)}.\hskip 1em plus 0.5em minus 0.4em\relax IEEE, 2023, pp. 214--219.

\bibitem[Mahanipour and Khamfroush(2024)]{mahanipour2024fmlfs}
------, ``Fmlfs: A federated multi-label feature selection based on information theory in iot environment,'' \emph{arXiv preprint arXiv:2405.00524}, 2024.

\bibitem[Song et~al.(2024)Song, Cao, Miao, Yang, and Yu]{song2024causal}
Y.~Song, D.~Cao, J.~Miao, S.~Yang, and K.~Yu, ``Causal multi-label feature selection in federated setting,'' \emph{arXiv preprint arXiv:2403.06419}, 2024.

\bibitem[Dubois and Prade(1990)]{dubois1990rough}
D.~Dubois and H.~Prade, ``Rough fuzzy sets and fuzzy rough sets,'' \emph{International Journal of General System}, vol.~17, no. 2-3, pp. 191--209, 1990.

\bibitem[Wan et~al.(2021)Wan, Chen, Li, Yuan, Liu, and Huang]{wan2021interactive}
J.~Wan, H.~Chen, T.~Li, Z.~Yuan, J.~Liu, and W.~Huang, ``Interactive and complementary feature selection via fuzzy multigranularity uncertainty measures,'' \emph{IEEE Transactions on Cybernetics}, vol.~53, no.~2, pp. 1208--1221, 2021.

\bibitem[Luo et~al.(2016)Luo, Gong, Hu, Duan, and Ma]{luo2016ensemble}
D.~Luo, C.~Gong, R.~Hu, L.~Duan, and S.~Ma, ``Ensemble enabled weighted pagerank,'' \emph{arXiv preprint arXiv:1604.05462}, 2016.

\end{thebibliography}
\end{document}